\documentclass[journal]{IEEEtran}

\usepackage[utf8]{inputenc}
\usepackage[T1]{fontenc}
\usepackage{amsmath,amsfonts}
\usepackage{graphicx}
\usepackage{xcolor}
\usepackage{booktabs}
\usepackage{multirow}
\usepackage{array}
\usepackage{colortbl}
\usepackage{url}
\usepackage{cite}
\usepackage{pifont}
\usepackage[ruled,vlined,linesnumbered]{algorithm2e}

\graphicspath{{IMG/}{Consistent-Inversion/IMG/}}

\newcommand{\source}{\mathcal{S}}
\newcommand{\target}{\mathcal{T}}
\newcommand{\emptycond}{\emptyset}
\definecolor{oursblue}{HTML}{EAF2FB}
\definecolor{accentblue}{HTML}{2F6FBB}
\newcommand{\best}[1]{\textbf{#1}}
\newcommand{\oursrow}{\rowcolor{oursblue}}

\begin{document}

\title{Consistent-Inversion: Reverse Consistency Guidance\\for Structure-Preserving Visual Editing}

\author{Xiaocheng~Lu,
        Jingcai~Guo,~\IEEEmembership{Senior~Member,~IEEE},
        and~Song~Guo,~\IEEEmembership{Fellow,~IEEE}%
\thanks{This work has been submitted to the IEEE for possible publication. Copyright may be transferred without notice, after which this version may no longer be accessible.}
\thanks{Xiaocheng Lu and Song Guo are with the Hong Kong University of Science and Technology, Hong Kong SAR, China. E-mails: xiaochenglu1997@gmail.com; songguo@ust.hk.}
\thanks{Jingcai Guo is with The Hong Kong Polytechnic University, Hong Kong SAR, China. E-mail: jc-jingcai.guo@polyu.edu.hk.}
\thanks{Corresponding author: Song Guo (songguo@ust.hk).}
}

\markboth{IEEE Transactions on Multimedia,~Vol.~XX, No.~XX, 2026}%
{Lu \MakeLowercase{\textit{et al.}}: Consistent-Inversion}

\maketitle

\begin{abstract}
Text-guided diffusion models have become effective tools for real-image visual editing, where the edited image must follow a target instruction while preserving editing-irrelevant structure. Most training-free editors rely on inversion: a source image is mapped to a noisy latent trajectory and the terminal latent is reused for target-prompt denoising. This reuse is useful for preservation, but it also couples source reconstruction and target editing. The resulting trajectory mismatch may either damage background/layout details or over-constrain the intended edit.
This paper presents Consistent-Inversion, a training-free reverse consistency guidance framework for structure-preserving visual editing. Instead of treating the inverted source latent as a fixed initialization, Consistent-Inversion checks whether an intermediate target trajectory can be reversed toward the source inversion trajectory under the source prompt. To make this check well-defined, we construct an auxiliary target-side noise representation, perform source-guided reverse denoising, and use the resulting reverse consistency discrepancy as a correction signal for selected early target denoising steps. The method does not update model parameters, is compatible with inversion-based editors, and introduces only a small inference overhead when applied sparsely. Experiments on PIE-Bench show that Consistent-Inversion improves background and structural fidelity under a unified SD3.5 protocol while maintaining target-prompt alignment, and compatibility experiments further verify the same correction principle on classical Stable-Diffusion inversion pipelines.
\end{abstract}


\section{Introduction}

Text-guided visual editing modifies objects, attributes, styles, or backgrounds according to natural-language instructions while preserving content that should remain unchanged. It is an important capability for multimedia content production, media post-processing, and interactive image manipulation systems. Diffusion models provide strong realism and semantic controllability, making them attractive backbones for text-based editing~\cite{ho2020denoising,song2020denoising,brooks2023instructpix2pix}. However, real-image editing differs from text-to-image generation: the editor must first recover a latent trajectory for the source image and then guide that trajectory toward the target prompt without damaging editing-irrelevant structure.

A widely used training-free solution is inversion-based editing. Given a source image and a source prompt, DDIM inversion maps the image from $z_0$ to a noisy latent $z_T$~\cite{song2020denoising}. The target prompt is then used during reverse denoising to synthesize the edited image. Existing methods improve this pipeline through attention replacement~\cite{hertz2022prompt,cao2023masactrl,parmar2023zero}, latent replacement~\cite{meng2021sdedit,avrahami2022blended,couairon2022diffedit}, prompt optimization~\cite{mokady2023null,dong2023prompt}, or exact/coupled inversion~\cite{wallace2023edict}. These techniques have made training-free editing practical, but they still inherit a trajectory mismatch: the same inverted source latent is used both as a source reconstruction state and as the initial state of target editing.

This mismatch matters because the noised source latent is not pure Gaussian noise. It contains source-specific structural cues, especially at the early denoising stages. Reusing it helps preserve layout, but it can also bias the target branch toward the source distribution. Conversely, aggressive target guidance may destroy background details and global structure. Current training-free editors usually control this tradeoff with guidance scale, attention injection windows, or manually selected latent replacement steps. These controls are useful but indirect: they do not explicitly ask whether the edited trajectory is still consistent with the source trajectory.

We propose Consistent-Inversion, a reverse consistency guidance method for correcting this mismatch. Our central observation is that a structurally reliable target edit should remain approximately reversible with respect to the source condition. If an intermediate target state is mapped back to a source-guided trajectory, it should return near the corresponding source inversion state. The discrepancy between this reversed state and the source trajectory estimates structural drift in the target branch. We construct an auxiliary target-side noise representation, compute the reverse consistency discrepancy, and inject the resulting offset into selected early denoising steps. Since global layout is largely determined early in denoising, this selected-timestep design captures the main preservation benefit while keeping the additional cost limited.

\begin{figure*}[t]
  \centering
  \includegraphics[width=\textwidth]{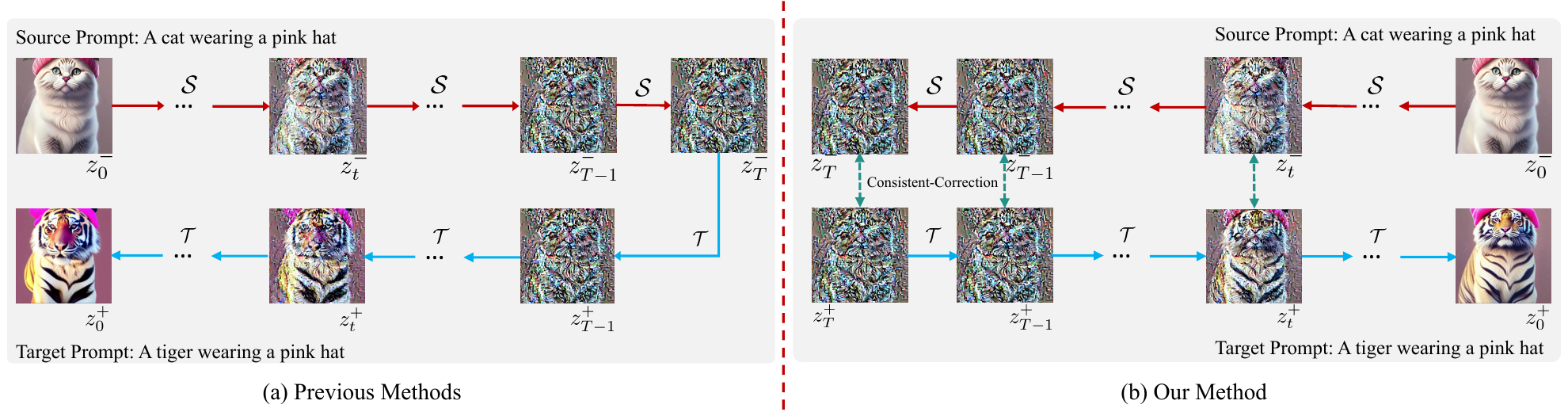}
  \caption{Motivation and comparison with conventional inversion-based editing. The upper branch denotes DDIM inversion of the source image, and the lower branch denotes target-prompt denoising. Existing methods mainly preserve source content through attention or latent reuse. Consistent-Inversion additionally checks whether the target branch can be reversed toward the source trajectory and uses the resulting discrepancy to correct early denoising.}
  \label{fig:motivation}
\end{figure*}

For visual editing systems, efficiency, controllability, and compatibility are critical. Consistent-Inversion is training-free, does not optimize model parameters for each image, and can be attached to inversion-based editors such as Prompt-to-Prompt, Plug-and-Play, and Direct Inversion. It also provides an interpretable correction signal, making the preservation mechanism easier to diagnose than purely heuristic guidance schedules.

The contributions are summarized as follows:
\begin{itemize}
    \item We identify a trajectory mismatch in inversion-based diffusion editing: the source inversion trajectory and the target editing trajectory are coupled through the same noised source latent, which can cause structural drift or over-preservation.
    \item We propose a bidirectional reverse consistency correction that estimates editing-induced structural discrepancy by source-guided reverse denoising from an auxiliary target-side noise representation.
    \item We develop an efficient selected-timestep implementation that improves structure preservation with limited additional inference cost and remains compatible with existing training-free editing pipelines.
    \item We provide a unified SD3.5 evaluation, classical-backbone compatibility results, ablation studies, and runtime analysis demonstrating improved preservation-fidelity tradeoffs.
\end{itemize}

\section{Related Work}

\subsection{Training-Based and Optimization-Based Editing}

Diffusion models have achieved strong performance in image generation and editing~\cite{ho2020denoising,song2020denoising,dhariwal2021diffusion}. Text-guided editing methods can be roughly grouped into training-based, test-time optimization, and training-free approaches. Training-based methods learn editing behavior from supervised or weakly supervised data. InstructPix2Pix~\cite{brooks2023instructpix2pix}, SmartEdit~\cite{huang2023smartedit}, and instruction-oriented editing models directly learn to follow textual commands. Domain-specific or attribute-driven methods, such as DiffusionCLIP~\cite{kim2022diffusionclip} and StyleDiffusion~\cite{wang2023stylediffusion}, provide stronger control in constrained domains such as face or style manipulation. These methods are powerful when training data is available, but they require model training and may not transfer well to arbitrary real images or open-ended prompts.

Test-time optimization methods adapt the model, text embedding, or latent representation to a particular source image~\cite{kawar2023imagic,mokady2023null,dong2023prompt,shi2023dragdiffusion}. Imagic~\cite{kawar2023imagic} optimizes image-specific representations and model parameters. Null-text inversion~\cite{mokady2023null} optimizes unconditional embeddings to improve reconstruction before editing. Prompt tuning inversion~\cite{dong2023prompt} similarly improves alignment by adapting prompt representations. DragDiffusion~\cite{shi2023dragdiffusion} optimizes latent variables for point-based manipulation. Such methods often improve reconstruction and controllability, but their per-image optimization cost limits interactive use.

\subsection{Training-Free Attention and Feature Control}

Training-free editing avoids model tuning and instead manipulates the denoising process. Prompt-to-Prompt~\cite{hertz2022prompt} shows that cross-attention maps can control the spatial binding between prompt tokens and generated content. Plug-and-Play~\cite{tumanyan2023plug} injects intermediate diffusion features to preserve source structure. MasaCtrl~\cite{cao2023masactrl} uses mutual self-attention control to improve consistency in image synthesis and editing. ProxEdit~\cite{han2024proxedit} introduces proximal guidance for tuning-free real-image editing, and LEDITS++~\cite{brack2024leditspp} improves textual image manipulation with efficient inversion and implicit masking. These methods show that internal diffusion features contain rich layout information. However, attention and feature controls are usually indirect: they preserve selected representations but do not explicitly verify whether the target trajectory remains consistent with the source trajectory.

\subsection{Inversion and Trajectory Control}

DDIM inversion~\cite{song2020denoising} maps a real image into the latent space of a deterministic diffusion sampler. However, the inversion and denoising trajectories are only approximately reversible for finite timesteps and conditional diffusion models. Recent work has improved inversion through exact coupled transformations~\cite{wallace2023edict}, negative-prompt guidance~\cite{miyake2023negative}, prompt tuning~\cite{dong2023prompt}, source trajectory reuse~\cite{ju2023direct}, iterative noising~\cite{garibi2024renoise}, and region- or time-dependent stochasticity~\cite{kang2024eta}. Tight Inversion~\cite{kadosh2025tight} studies image-conditioned inversion, and DCI~\cite{li2025dci} jointly conditions on source prompt and reference image to improve reconstruction and editability. These newer methods make the experimental comparison more demanding than earlier DDIM/NTI-based baselines. Our method takes a different view: instead of only improving reconstruction or initial noise quality, we use the discrepancy between source-guided reverse editing and the source inversion path as an online diagnostic signal for target-branch structural drift.

\subsection{Recent Flow and Few-Step Editing}

Recent generative backbones have shifted from classical multi-step diffusion models toward rectified flows and one-step/few-step models. RF Inversion~\cite{rout2024rfinversion} formulates inversion and editing for rectified flow models through stochastic differential equations and dynamic optimal control. FlashEdit~\cite{wu2025flashedit} emphasizes real-time localized editing through one-step inversion and editing. SteerFlow~\cite{dao2026steerflow} studies faithful flow-based trajectory steering with adaptive masking. TFinv~\cite{wu2026tfinv} studies training-free inversion for one-step diffusion models and reports strong PIE-Bench efficiency. These methods highlight a broader trend: modern visual editing increasingly depends on efficient trajectories, accurate inversion, and controllable correction. Consistent-Inversion is complementary to this trend because it improves an existing trajectory rather than requiring additional model training.

\subsection{Structure Preservation and Consistency}

Structure preservation is often handled by attention replacement, feature injection, masks, or latent blending~\cite{hertz2022prompt,avrahami2022blended,couairon2022diffedit,cao2023masactrl}. DiffEdit~\cite{couairon2022diffedit} estimates masks to localize semantic changes, while Blended Diffusion~\cite{avrahami2022blended} combines masked image regions during diffusion. These methods preserve background regions effectively when masks are accurate, but they may struggle when edit-relevant and edit-irrelevant structures are entangled. Consistent-Inversion is complementary: it estimates preservation at the trajectory level, where both background layout and object posture are encoded in intermediate latents. The resulting offset can be combined with mask or attention controls rather than replacing them.

\section{Preliminaries}

\subsection{Problem Formulation}

Given a source image $x_s$, a source prompt $\source$, and a target prompt $\target$, text-guided real-image editing synthesizes an edited image $x_t$ that satisfies two requirements. First, $x_t$ should align with the target prompt in the edited region. Second, content irrelevant to the edit should remain close to $x_s$. With an optional editing mask $M$, this tradeoff can be conceptually written as
\begin{equation}
\label{eq:editing_objective}
\min_{x_t} \ 
\mathcal{D}_{\mathrm{edit}}(M\odot x_t,\target)
+\beta \mathcal{D}_{\mathrm{pres}}((1-M)\odot x_t,(1-M)\odot x_s),
\end{equation}
where $\mathcal{D}_{\mathrm{edit}}$ measures target alignment, $\mathcal{D}_{\mathrm{pres}}$ measures source preservation, and $\beta$ controls their tradeoff. Training-free diffusion editors do not directly optimize Eq.~\eqref{eq:editing_objective}; instead, they manipulate the latent denoising trajectory through prompt guidance, attention control, feature injection, or inversion trajectory reuse. Consistent-Inversion can be interpreted as adding an implicit trajectory-level preservation constraint: the edited latent should remain reversible toward the source inversion trajectory under the source condition.

\subsection{DDIM Sampling and Inversion}

Let $z_t$ denote the latent at timestep $t$, $\alpha_t$ the noise schedule, and $\epsilon_\theta(z_t,t,\mathcal{C})$ the noise predicted by a pretrained conditional diffusion model under condition $\mathcal{C}$. DDIM sampling generates $z_{t-1}$ from $z_t$ as
\begin{equation}
\label{eq:ddim_sampling}
z_{t-1}=\sqrt{\frac{\alpha_{t-1}}{\alpha_t}}z_t+
\left(\sqrt{\frac{1}{\alpha_{t-1}}-1}-\sqrt{\frac{1}{\alpha_t}-1}\right)
\epsilon_\theta(z_t,t,\mathcal{C}).
\end{equation}
DDIM inversion approximates the reverse mapping from $z_t$ to $z_{t+1}$:
\begin{equation}
\label{eq:ddim_inversion}
z_{t+1}=\sqrt{\frac{\alpha_{t+1}}{\alpha_t}}z_t+
\left(\sqrt{\frac{1}{\alpha_{t+1}}-1}-\sqrt{\frac{1}{\alpha_t}-1}\right)
\epsilon_\theta(z_t,t,\mathcal{C}).
\end{equation}

\subsection{Classifier-Free Guidance}

For text-guided diffusion models, classifier-free guidance combines conditional and unconditional predictions:
\begin{equation}
\label{eq:cfg}
\epsilon_\theta(z_t,t,\mathcal{C},\emptycond)
=w\epsilon_\theta(z_t,t,\mathcal{C})+(1-w)\epsilon_\theta(z_t,t,\emptycond),
\end{equation}
where $w$ is the guidance scale and $\emptycond$ denotes the null condition.

\section{Method}

\subsection{Trajectory Mismatch in Inversion-Based Editing}

Let $\source$ and $\target$ denote the source and target prompts. In a standard inversion-based editing pipeline, the source image latent $z^-_0$ is inverted under $\source$:
\begin{equation}
\label{eq:source_inversion}
z^-_{t+1}=\mathrm{Inv}(z^-_t,t,\source),
\end{equation}
where $\mathrm{Inv}(\cdot)$ denotes Eq.~\eqref{eq:ddim_inversion}. Editing then starts from $z^-_T$ and denoises under $\target$:
\begin{equation}
\label{eq:target_sampling}
z^+_{t-1}=\mathrm{Sam}(z^+_t,t,\target),
\quad z^+_T=z^-_T,
\end{equation}
where $\mathrm{Sam}(\cdot)$ denotes Eq.~\eqref{eq:ddim_sampling}. This formulation couples two different goals. The source branch seeks faithful reconstruction, while the target branch seeks semantic modification. Because $z^-_T$ still carries source information, target denoising may over-preserve source semantics. Conversely, when the target condition dominates, editing-irrelevant background and layout may drift. The key question is therefore how to preserve the source-consistent part of the trajectory without preventing the intended edit.

\begin{figure*}[t]
  \centering
  \includegraphics[width=\textwidth]{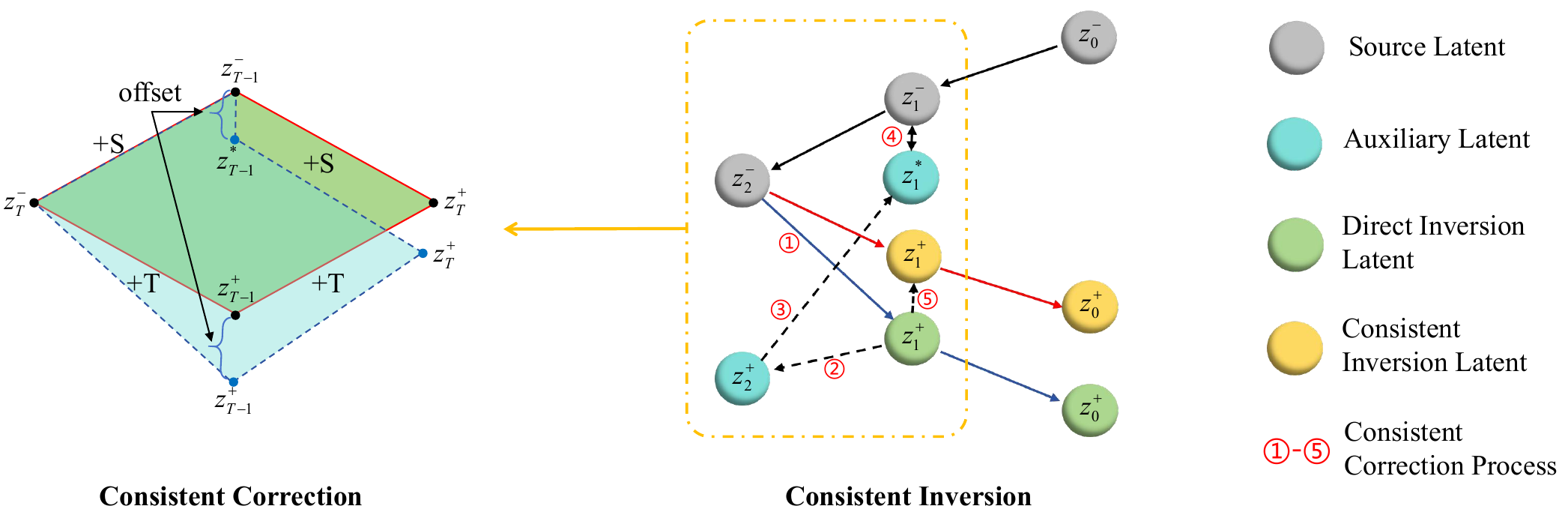}
  \caption{Overall pipeline of Consistent-Inversion. The source trajectory $z^-_t$ is obtained by DDIM inversion under the source prompt. The target trajectory $z^+_t$ is denoised under the target prompt. At selected early timesteps, an auxiliary target-side noise representation is constructed and source-guided reverse denoising estimates the reverse consistency discrepancy, which is then injected into the target branch as a correction offset.}
  \label{fig:pipeline}
\end{figure*}

\subsection{Bidirectional Reverse Consistency}

We estimate structural drift by asking whether the current target state can be reversed back to the source trajectory. Consider an early denoising step that produces $z^+_{k}$ from $z^+_{k+1}$. We first construct an auxiliary target-side noise representation by inverting the current target state back toward $T$ under the target prompt:
\begin{equation}
\label{eq:auxiliary_target}
\hat{z}^+_{T}=\mathrm{Inv}_{k\rightarrow T}(z^+_{k},\target).
\end{equation}
Starting from $\hat{z}^+_{T}$, we then perform source-guided denoising back to timestep $k$:
\begin{equation}
\label{eq:reverse_source}
z^*_{k}=\mathrm{Sam}_{T\rightarrow k}(\hat{z}^+_{T},\source).
\end{equation}
If the target edit preserves essential source structure, $z^*_k$ should be close to the source inversion state $z^-_k$. We define the reverse consistency offset as
\begin{equation}
\label{eq:offset}
O^-_k=z^-_k-z^*_k.
\end{equation}
The corrected target latent is then
\begin{equation}
\label{eq:correction}
z^+_k \leftarrow z^+_k+\lambda_k O^-_k,
\end{equation}
where $\lambda_k$ controls the correction strength.

\subsection{Step-Wise Construction of Consistent Correction}

We now instantiate the above idea at a single early denoising step and then extend it to multiple selected timesteps. For clarity, consider the first target denoising step from $T$ to $T-1$. The target prompt first guides the source-side terminal latent $z^-_T$ into a target-side intermediate latent:
\begin{equation}
\label{eq:first_target_guidance}
\begin{aligned}
z^+_{T-1}
&=\sqrt{\frac{\alpha_{T-1}}{\alpha_T}}z^-_T \\
&\quad+
\left(\sqrt{\frac{1}{\alpha_{T-1}}-1}-\sqrt{\frac{1}{\alpha_T}-1}\right)
\epsilon_\theta(z^-_T,T,\target,\emptycond).
\end{aligned}
\end{equation}
In a conventional editor, $z^+_{T-1}$ is directly passed to the following denoising steps. Consistent-Inversion instead evaluates whether this newly edited latent remains reversible under the source condition. Since $z^+_{T-1}$ is not naturally paired with a known terminal target latent, we first construct an auxiliary target-side noise representation:
\begin{equation}
\label{eq:step_aux}
\begin{aligned}
\hat{z}^+_T
&=\sqrt{\frac{\alpha_T}{\alpha_{T-1}}}z^+_{T-1}\\
&\quad+
\left(\sqrt{\frac{1}{\alpha_T}-1}-\sqrt{\frac{1}{\alpha_{T-1}}-1}\right)
\epsilon_\theta(z^+_{T-1},T-1,\target).
\end{aligned}
\end{equation}
Then we reverse-denoise this auxiliary representation with the source prompt:
\begin{equation}
\label{eq:step_source_reverse}
\begin{aligned}
z^*_{T-1}
&=\sqrt{\frac{\alpha_{T-1}}{\alpha_T}}\hat{z}^+_T\\
&\quad+
\left(\sqrt{\frac{1}{\alpha_{T-1}}-1}-\sqrt{\frac{1}{\alpha_T}-1}\right)
\epsilon_\theta(\hat{z}^+_T,T,\source,\emptycond).
\end{aligned}
\end{equation}
The source trajectory already contains $z^-_{T-1}$ from DDIM inversion. Therefore, the reverse consistency offset at this step is
\begin{equation}
\label{eq:step_offset}
O^-_{T-1}=z^-_{T-1}-z^*_{T-1}.
\end{equation}
Finally, the target latent is corrected by
\begin{equation}
\label{eq:step_refine}
z^+_{T-1}\leftarrow z^+_{T-1}+\lambda_{T-1}O^-_{T-1}.
\end{equation}
The same construction applies to any selected timestep $k$. When $k<T-1$, the method inverts $z^+_k$ back to the terminal auxiliary representation and source-denoises it back to $k$. This is more expensive than a one-step check, which motivates our selected-timestep implementation.

\subsection{Efficient Selected-Timestep Correction}

Computing Eq.~\eqref{eq:auxiliary_target} and Eq.~\eqref{eq:reverse_source} at every timestep is unnecessary and inefficient. Diffusion denoising tends to establish global layout and low-frequency structure in earlier stages, while later stages refine local appearance and texture. We therefore apply correction only at a small set of early timesteps $\mathcal{K}$:
\begin{equation}
\label{eq:selected_steps}
\mathcal{K}=\{T-\Delta_1,T-\Delta_2,\ldots,T-\Delta_m\}, \quad m \ll T.
\end{equation}
This selected-timestep strategy preserves the main structural benefit while keeping runtime close to the underlying editor.

\subsection{Computational Cost}

Let $T$ be the number of DDIM sampling steps and $m=|\mathcal{K}|$ the number of corrected timesteps. A standard inversion-based editor requires one source inversion pass and one target denoising pass, giving approximately $2T$ denoising-network evaluations, excluding editor-specific attention or feature operations. Consistent-Inversion adds one auxiliary inversion path and one source-guided reverse path for each corrected timestep. Dense correction would be expensive, but the selected-timestep configuration keeps $m$ small and places corrections near the beginning of denoising, where structural information is most influential. Empirically, the default setting increases single-image runtime only slightly over Direct Inversion while improving reverse structural consistency, as shown in Sec.~\ref{sec:runtime}.

This design is suitable for practical visual editing systems, where users often prefer a controllable quality-speed tradeoff rather than uniformly expensive correction at every timestep. The correction set $\mathcal{K}$ and scale schedule $\lambda_k$ can be configured according to latency budget, desired preservation strength, and availability of editing masks.

\begin{algorithm}[t]
\caption{Consistent-Inversion}
\label{alg:consistent_inversion}
\KwIn{Source latent $z^-_0$, source prompt $\source$, target prompt $\target$, timesteps $T$, correction set $\mathcal{K}$, scale schedule $\lambda_k$}
\KwOut{Edited latent $z^+_0$}
\For{$t=0$ \KwTo $T-1$}{
    $z^-_{t+1}\leftarrow \mathrm{Inv}(z^-_t,t,\source)$\;
}
$z^+_T\leftarrow z^-_T$\;
\For{$t=T$ \KwTo $1$}{
    $z^+_{t-1}\leftarrow \mathrm{Sam}(z^+_t,t,\target)$\;
    \If{$t-1\in\mathcal{K}$}{
        $\hat{z}^+_T\leftarrow \mathrm{Inv}_{t-1\rightarrow T}(z^+_{t-1},\target)$\;
        $z^*_{t-1}\leftarrow \mathrm{Sam}_{T\rightarrow t-1}(\hat{z}^+_T,\source)$\;
        $O^-_{t-1}\leftarrow z^-_{t-1}-z^*_{t-1}$\;
        $z^+_{t-1}\leftarrow z^+_{t-1}+\lambda_{t-1}O^-_{t-1}$\;
    }
}
\end{algorithm}

\subsection{Compatibility with Existing Editors}

Consistent-Inversion operates at the latent trajectory level and does not alter the pretrained diffusion network. It can therefore be combined with attention-based editors such as Prompt-to-Prompt~\cite{hertz2022prompt}, feature-injection editors such as Plug-and-Play~\cite{tumanyan2023plug}, and trajectory-based methods such as Direct Inversion~\cite{ju2023direct}. In our implementation, the base editor performs its original target denoising step, after which Consistent-Inversion applies the reverse consistency correction at selected timesteps.

\subsection{Design Rationale}

The auxiliary target-side noise in Eq.~\eqref{eq:auxiliary_target} is necessary because $z^+_k$ and $z^-_k$ lie on different conditional trajectories. Directly subtracting them would mix target-guided semantic change with source-guided preservation error. By first mapping the target state to an auxiliary noise representation and then denoising it under the source prompt, Consistent-Inversion compares the two branches through the same DDIM geometry used by inversion and sampling. The resulting offset is therefore a prompt-aware trajectory discrepancy rather than a fixed pixel-space or latent-space residual.

The correction assumes that, during early denoising, reverse consistency discrepancy and target-branch structural drift have approximately aligned directions. This assumption is most reliable when source and target prompts share scene layout and differ mainly in object category, texture, color, material, or style. For edits involving large viewpoint changes, object count changes, or global spatial rearrangement, the correction should be applied more conservatively. This observation motivates sparse early-timestep correction with moderate scales rather than exact cycle consistency throughout the whole trajectory.

\section{Experiments}

\subsection{Experimental Setup}

We evaluate Consistent-Inversion on PIE-Bench~\cite{ju2023direct}, which contains 700 images and ten prompt-based editing types: random editing, object change, object addition, object deletion, object content modification, object pose alteration, object color change, object material change, background change, and image style change. Each image is associated with a source prompt, a target prompt, an editing instruction, editing subjects, and an editing mask. This benchmark is suitable for our study because it explicitly measures both target edit fidelity and preservation of editing-irrelevant regions.

We organize the experiments into two protocols. The primary protocol uses a unified SD3.5 setting on the full PIE-Bench benchmark. In this setting, all rows in the main table are evaluated with the same model family, prompt set, image set, and metric code, so the comparison isolates the effect of the inversion or correction strategy. The secondary protocol follows the classical Stable-Diffusion-v1.4, Prompt-to-Prompt, and 50-step DDIM setting used by Direct Inversion~\cite{ju2023direct}. We use this protocol only as compatibility evidence for earlier inversion backbones, not as a direct ranking against SD3.5 results.

Unless otherwise stated, the default correction scale is 0.5, and the number of corrected timesteps is set to a small fraction of the denoising trajectory to balance preservation and efficiency. The correction is applied at early timesteps, where latent trajectories mainly determine global layout and editing-irrelevant structure.

\subsection{Evaluation Metrics}

Following prior work, we report background LPIPS~\cite{zhang2018unreasonable}, DINO-ViT structure distance, LPIPS, MSE, PSNR, SSIM~\cite{wang2004image}, and CLIP similarity. Background LPIPS and DINO-ViT distance measure preservation of editing-irrelevant regions and semantic layout. LPIPS, MSE, PSNR, and SSIM evaluate reconstruction-level fidelity. CLIP image-image similarity measures source-image preservation, while CLIP text-image and directional CLIP similarities evaluate target-prompt alignment. For runtime analysis, we measure wall-clock inference time per image on a single GPU.

We also report reverse SSIM (R-SSIM), a cycle-inspired consistency metric. Given an edited image $x_t$, we reverse-edit it back to the source prompt and obtain $\tilde{x}_s$. R-SSIM is defined as
\begin{equation}
\label{eq:rssim}
\mathrm{R\text{-}SSIM}=\mathrm{SSIM}(\tilde{x}_s,x_s).
\end{equation}
This metric does not replace target-prompt fidelity metrics; instead, it measures whether the edited result still contains enough source-consistent structure to be reversed back. A higher R-SSIM indicates better trajectory-level preservation, which directly matches the motivation of Consistent-Inversion. Since a structure-preserving editor should not be judged by a single scalar, we make paired claims throughout the experiments: preservation-sensitive metrics should improve while CLIP-based prompt alignment remains close to the underlying editor.

\subsection{Per-Type SD3.5 Analysis}

PIE-Bench mixes local object edits, attribute edits, background changes, and global style edits, whose annotated edit regions range from small object-deletion masks to nearly full-image style changes. To address whether the average gain in Table~\ref{tab:main_results} comes from only a few easy categories, we aggregate the full 700-image SD3.5 results by editing type. Table~\ref{tab:per_type} compares Direct Inversion and Consistent-Inversion on representative preservation and editability metrics. Consistent-Inversion improves BG-LPIPS in nine out of ten categories and LPIPS in all ten categories. The only exception on BG-LPIPS is image style change, where background-only metrics become less diagnostic; even in this category, LPIPS still improves from 0.4647 to 0.4292.

The per-type results support the intended behavior of the method. Larger gains appear for color change and background change, where target semantics often leak into surrounding texture and illumination. Local deletion and object change also benefit, indicating that the correction helps keep unedited context stable. CLIP-T remains close to Direct Inversion across categories, with a slight improvement for object deletion. This confirms that the method mainly reduces editing-irrelevant drift rather than obtaining preservation by suppressing the target edit.

\begin{table*}[t]
\centering
\caption{Per-type SD3.5 comparison on PIE-Bench. D denotes Direct Inversion and O denotes Direct Inversion + Consistent-Inversion.}
\label{tab:per_type}
\setlength{\tabcolsep}{3.0pt}
\renewcommand{\arraystretch}{1.08}
\scriptsize
\begin{tabular}{@{}lccccccc@{}}
\toprule
\multirow{2}{*}{Editing type} & \multirow{2}{*}{Images} & \multicolumn{2}{c}{BG-LPIPS$\downarrow$} & \multicolumn{2}{c}{LPIPS$\downarrow$} & \multicolumn{2}{c}{CLIP-T$\uparrow$} \\
\cmidrule(lr){3-4}\cmidrule(lr){5-6}\cmidrule(l){7-8}
& & D & O & D & O & D & O \\
\midrule
Random editing & 140 & 0.2348 & \best{0.2182} & 0.4562 & \best{0.4239} & 0.3244 & 0.3244 \\
Object change & 80 & 0.3114 & \best{0.2909} & 0.4406 & \best{0.4193} & 0.3244 & 0.3238 \\
Object addition & 80 & 0.1410 & \best{0.1345} & 0.3725 & \best{0.3493} & 0.3309 & 0.3303 \\
Object deletion & 80 & 0.3390 & \best{0.3170} & 0.4342 & \best{0.4106} & 0.3153 & \best{0.3161} \\
Content modification & 40 & 0.2401 & \best{0.2312} & 0.3899 & \best{0.3760} & 0.3186 & 0.3159 \\
Pose alteration & 40 & 0.2354 & \best{0.2259} & 0.3739 & \best{0.3616} & 0.3249 & 0.3245 \\
Color change & 40 & 0.2996 & \best{0.2722} & 0.4446 & \best{0.4071} & 0.3336 & 0.3303 \\
Material change & 40 & 0.2718 & \best{0.2557} & 0.4295 & \best{0.4099} & 0.3380 & 0.3335 \\
Background change & 80 & 0.1874 & \best{0.1708} & 0.5290 & \best{0.4792} & 0.3282 & 0.3277 \\
Style change & 80 & \best{0.0069} & 0.0071 & 0.4647 & \best{0.4292} & 0.3440 & 0.3405 \\
\midrule
All & 700 & 0.2194 & \best{0.2051} & 0.4409 & \best{0.4122} & \best{0.3278} & 0.3267 \\
\bottomrule
\end{tabular}
\end{table*}

\subsection{Main Comparison on PIE-Bench}

Table~\ref{tab:main_results} reports the main SD3.5 comparison on the full 700-image PIE-Bench protocol. Because recent editors often use different backbones, samplers, acceleration strategies, and code assumptions, we avoid mixing incompatible settings into one leaderboard. Instead, the main table uses a unified SD3.5 setting and evaluates all entries with the same PIE-Bench metric code. Euler Flow Inversion corresponds to DDIM-style inversion, Negative-Prompt Flow Inversion corresponds to NPI, and Direct Inversion denotes the direct trajectory-reuse baseline. We then insert the proposed reverse consistency correction into Direct Inversion.

The simple flow-inversion controls obtain competitive CLIP text or directional scores, but they substantially degrade preservation. Direct Inversion is the strongest of these controls, and adding reverse consistency correction consistently improves preservation metrics: BGLPIPS decreases from 0.2194 to 0.2051, DINO distance from 0.0656 to 0.0631, LPIPS from 0.4409 to 0.4122, and MSE from 0.03859 to 0.03062. Reconstruction-oriented metrics show the same trend, with PSNR increasing from 15.19 dB to 16.06 dB, SSIM from 0.6318 to 0.6578, and CLIP image similarity from 0.8375 to 0.8437. The CLIP text and directional scores remain close to the Direct Inversion baseline, indicating that the correction improves structure preservation without simply suppressing the edit.

Since many PIE-Bench edits occupy only a modest fraction of the image, unnecessary changes outside the target region are heavily penalized by BG-LPIPS, DINO distance, and reconstruction metrics. Direct Inversion already reduces this drift compared with generic flow-inversion controls, but it still reuses a forward trajectory that may not be reversible under the source prompt after target-conditioned denoising. Consistent-Inversion directly targets this failure mode, so its largest relative gain appears on MSE and LPIPS rather than on CLIP-T. This is the desired behavior: the method should reduce editing-irrelevant distortion while preserving the original editor's semantic change.

For completeness, Table~\ref{tab:sd14_compat} keeps the original SD-v1.4 comparison against DDIM Inversion, NPI, ProxNPI, and Direct Inversion. We report it separately because those methods use a different diffusion backbone and DDIM-style sampler; mixing them with SD3.5 rows in a single ranking would conflate method improvements with model changes. This separation lets the main table support the current-backbone comparison while the classical table demonstrates that the same correction mechanism also benefits earlier inversion pipelines.

\begin{table*}[t]
\centering
\caption{Unified SD3.5 comparison on the full PIE-Bench benchmark. Ours inserts reverse consistency correction into Direct Inversion.}
\label{tab:main_results}
\setlength{\tabcolsep}{2.4pt}
\renewcommand{\arraystretch}{1.08}
\scriptsize
\begin{tabular}{@{}lccccccccc@{}}
\toprule
\multirow{2}{*}{Method} & \multicolumn{4}{c}{Preservation / structure} & \multicolumn{2}{c}{Background fidelity} & \multicolumn{3}{c}{CLIP alignment} \\
\cmidrule(lr){2-5}\cmidrule(lr){6-7}\cmidrule(l){8-10}
& BG-LPIPS$\downarrow$ & DINO$\downarrow$ & LPIPS$\downarrow$ & MSE$\downarrow$ & PSNR$\uparrow$ & SSIM$\uparrow$ & I$\uparrow$ & T$\uparrow$ & Dir$\uparrow$ \\
\midrule
Euler Flow Inv. (DDIM analogue) & 0.3241 & 0.0884 & 0.6015 & 0.08706 & 11.67 & 0.4857 & 0.8024 & 0.3266 & 0.0563 \\
Negative-Prompt Flow Inv. (NPI analogue) & 0.3706 & 0.0981 & 0.6747 & 0.11365 & 10.57 & 0.3981 & 0.7243 & 0.2963 & \textbf{0.0615} \\
Direct Inversion & 0.2194 & 0.0656 & 0.4409 & 0.03859 & 15.19 & 0.6318 & 0.8375 & \textbf{0.3278} & 0.0528 \\
\oursrow Direct Inversion + Consistent-Inversion & \best{0.2051} & \best{0.0631} & \best{0.4122} & \best{0.03062} & \best{16.06} & \best{0.6578} & \best{0.8437} & 0.3267 & 0.0515 \\
\bottomrule
\end{tabular}
\end{table*}

\begin{table*}[t]
\centering
\caption{Compatibility comparison under the classical SD-v1.4, Prompt-to-Prompt, and 50-step DDIM protocol.}
\label{tab:sd14_compat}
\setlength{\tabcolsep}{2.4pt}
\renewcommand{\arraystretch}{1.08}
\scriptsize
\begin{tabular}{@{}lccccccccc@{}}
\toprule
\multirow{2}{*}{Method} & \multicolumn{4}{c}{Preservation / structure} & \multicolumn{2}{c}{Background fidelity} & \multicolumn{3}{c}{CLIP alignment} \\
\cmidrule(lr){2-5}\cmidrule(lr){6-7}\cmidrule(l){8-10}
& BG-LPIPS$\downarrow$ & DINO$\downarrow$ & LPIPS$\downarrow$ & MSE$\downarrow$ & PSNR$\uparrow$ & SSIM$\uparrow$ & I$\uparrow$ & T$\uparrow$ & Dir$\uparrow$ \\
\midrule
DDIM Inv. & 0.2494 & 0.0699 & 0.4660 & 0.03472 & 15.11 & 0.5555 & 0.8452 & \textbf{0.3098} & 0.0201 \\
NPI~\cite{miyake2023negative} & 0.0833 & 0.0200 & 0.1939 & 0.00897 & 21.76 & 0.7185 & 0.9002 & 0.3047 & \textbf{0.0460} \\
ProxNPI~\cite{han2024proxedit} & 0.0786 & 0.0190 & 0.1781 & 0.00788 & 22.30 & 0.7294 & 0.9058 & 0.3031 & 0.0432 \\
Direct Inv.~\cite{ju2023direct} & 0.0633 & 0.0124 & 0.1581 & 0.00754 & 22.31 & 0.7367 & 0.9291 & 0.3092 & 0.0296 \\
\oursrow Ours-Direct-3 & 0.0602 & 0.0122 & 0.1501 & 0.00653 & 22.91 & 0.7435 & 0.9306 & 0.3089 & 0.0288 \\
\oursrow Ours-Direct-5 & 0.0584 & 0.0120 & 0.1449 & 0.00600 & 23.27 & 0.7475 & \textbf{0.9316} & 0.3083 & 0.0281 \\
\oursrow Ours-ProxNPI-3 & \best{0.0469} & \best{0.0101} & \best{0.1250} & \best{0.00567} & \best{23.98} & \best{0.7608} & 0.9277 & 0.3050 & 0.0329 \\
\bottomrule
\end{tabular}
\end{table*}

\subsection{Preservation-Editability Profile}

\begin{figure*}[t]
  \centering
  \includegraphics[width=.92\textwidth]{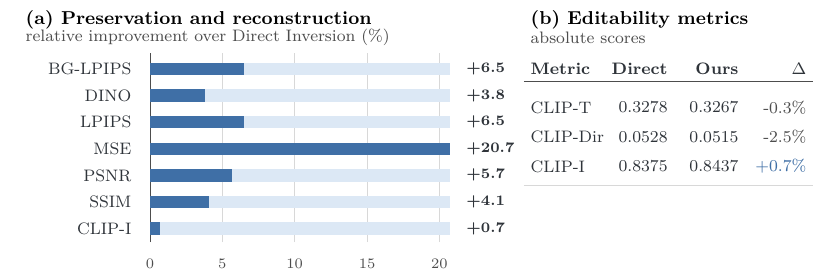}
  \caption{Relative SD3.5 preservation gains over Direct Inversion on the full PIE-Bench benchmark. Consistent-Inversion improves structure and background fidelity while keeping prompt-alignment metrics close to the base editor.}
  \label{fig:sd35_profile}
\end{figure*}

To make the preservation-editability tradeoff easier to inspect, Fig.~\ref{fig:sd35_profile} summarizes the relative SD3.5 gains of Consistent-Inversion over Direct Inversion. The largest improvements appear on background and reconstruction metrics, especially MSE, LPIPS, and BG-LPIPS. At the same time, CLIP-T and CLIP-Dir remain close to the Direct Inversion baseline. This profile is consistent with the goal of reverse consistency correction: it should reduce structural drift without turning the target branch into a source reconstruction process.

The profile also helps distinguish our contribution from a simple prompt-strength or guidance-scale adjustment. Increasing guidance often improves target-prompt scores but can amplify geometric drift, texture leakage, or background changes. In contrast, the reverse consistency correction produces a preservation-heavy gain profile while leaving CLIP-T nearly unchanged and slightly improving CLIP-I. This suggests that the correction acts mainly on source-consistent trajectory geometry rather than on the semantic direction of the edit.

\subsection{Mechanism Ablation under SD3.5}

To verify that the improvement comes from bidirectional reverse consistency rather than from simply adding a residual latent offset, we conduct a mechanism ablation on the full 700-image SD3.5 benchmark. Table~\ref{tab:sd35_mechanism} compares Direct Inversion, a residual-only control using the same correction scale, and the proposed cycle-based reverse consistency correction. The residual-only control is almost indistinguishable from Direct Inversion across preservation and prompt-alignment metrics. In contrast, the cycle-based variant substantially reduces BG-LPIPS, DINO distance, LPIPS, and MSE, while improving PSNR and SSIM. This confirms that the auxiliary target-side noise representation and source-guided reverse pass are necessary for obtaining an informative correction direction.

\begin{table*}[t]
\centering
\caption{Mechanism ablation on the full 700-image SD3.5 benchmark. A residual-only offset does not provide the preservation gain obtained by the full cycle-based reverse consistency correction.}
\label{tab:sd35_mechanism}
\setlength{\tabcolsep}{3.2pt}
\renewcommand{\arraystretch}{1.08}
\scriptsize
\begin{tabular}{@{}lcccccccc@{}}
\toprule
\multirow{2}{*}{Variant} & \multirow{2}{*}{Images} & \multicolumn{4}{c}{Preservation / structure} & \multicolumn{2}{c}{Background fidelity} & CLIP \\
\cmidrule(lr){3-6}\cmidrule(lr){7-8}\cmidrule(l){9-9}
& & BG-LPIPS$\downarrow$ & DINO$\downarrow$ & LPIPS$\downarrow$ & MSE$\downarrow$ & PSNR$\uparrow$ & SSIM$\uparrow$ & T$\uparrow$ \\
\midrule
Direct Inversion & 700 & 0.2194 & 0.0656 & 0.4409 & 0.03859 & 15.19 & 0.6318 & \best{0.3278} \\
Residual-only, $s{=}0.3$ & 700 & 0.2194 & 0.0656 & 0.4409 & 0.03859 & 15.19 & 0.6318 & 0.3277 \\
\oursrow Cycle reverse consistency, $s{=}0.3$ & 700 & \best{0.2051} & \best{0.0631} & \best{0.4122} & \best{0.03062} & \best{16.06} & \best{0.6578} & 0.3267 \\
\bottomrule
\end{tabular}
\end{table*}

\subsection{Ablation Study}

We ablate the correction steps and correction scale in Table~\ref{tab:ablation}. Increasing the number of corrected timesteps generally improves background preservation, while overly many correction steps may slightly reduce edit fidelity. This supports the selected-timestep design: applying correction only to early structural stages gives a better efficiency-fidelity balance than dense correction.

When the correction scale is fixed at 0.5, increasing the number of corrected steps from 1 to 20 gradually improves structure distance, PSNR, LPIPS, MSE, and SSIM. However, CLIP similarity peaks around three correction steps and then slightly decreases. This confirms that excessive preservation can suppress target-prompt changes. When the number of corrected steps is fixed to one, increasing the scale generally improves preservation metrics up to a moderate level, but an overly large scale does not further improve CLIP similarity. We therefore use a moderate default scale and a small correction set in the main experiments.

\begin{table*}[!t]
\centering
\caption{Ablation on correction steps and correction scale with 20 denoising steps. The dagger marks the default setting used for balanced preservation and editability.}
\label{tab:ablation}
\setlength{\tabcolsep}{3.0pt}
\renewcommand{\arraystretch}{1.08}
\scriptsize
\begin{tabular}{@{}ccccccccc@{}}
\toprule
\multicolumn{2}{c}{Correction} & Structure & \multicolumn{4}{c}{Background preservation} & \multicolumn{2}{c}{CLIP similarity} \\
\cmidrule(lr){1-2}\cmidrule(lr){3-3}\cmidrule(lr){4-7}\cmidrule(l){8-9}
Steps & Scale & Dist.$_{\times10^3}\downarrow$ & PSNR$\uparrow$ & LPIPS$_{\times10^3}\downarrow$ & MSE$_{\times10^4}\downarrow$ & SSIM$_{\times10^2}\uparrow$ & Whole$\uparrow$ & Edited$\uparrow$ \\
\midrule
\multicolumn{9}{@{}l}{\textit{Varying correction steps with scale fixed to 0.5}} \\
0 & 0 & 10.6 & 27.49 & 51.80 & 30.62 & 84.94 & 24.73 & 21.75 \\
1 & 0.5 & 9.1 & 27.85 & 49.80 & 28.21 & 85.41 & 24.94 & 21.81 \\
\oursrow 3$^\dagger$ & 0.5 & 8.7 & 27.96 & 49.20 & 27.59 & 85.47 & \textbf{24.96} & \textbf{21.82} \\
5 & 0.5 & 8.5 & 28.01 & 48.90 & 27.33 & 85.49 & 24.93 & 21.79 \\
10 & 0.5 & 8.4 & 28.08 & 48.50 & 27.03 & 85.52 & 24.92 & 21.77 \\
20 & 0.5 & \textbf{8.2} & \textbf{28.15} & \textbf{47.90} & \textbf{26.67} & \textbf{85.57} & 24.88 & 21.74 \\
\addlinespace[1pt]
\midrule
\multicolumn{9}{@{}l}{\textit{Varying correction scale with one corrected step}} \\
1 & 0.1 & 9.8 & 27.65 & 50.93 & 29.61 & 85.29 & \textbf{24.95} & 21.80 \\
1 & 0.3 & 9.5 & 27.76 & 50.40 & 28.82 & 85.35 & 24.94 & 21.79 \\
1 & 0.8 & \textbf{8.8} & 27.94 & 49.43 & \textbf{27.70} & 85.47 & \textbf{24.95} & \textbf{21.81} \\
1 & 1.0 & \textbf{8.8} & \textbf{27.96} & \textbf{49.33} & 27.75 & \textbf{85.48} & 24.92 & 21.79 \\
\bottomrule
\end{tabular}
\end{table*}

\subsection{Backbone Compatibility}

We further test whether the correction is tied to Direct Inversion. Table~\ref{tab:compact_backbone} inserts the same consistency correction into two classical inversion backbones while keeping Prompt-to-Prompt fixed. The paired gains on both Direct Inversion and ProxNPI show that Consistent-Inversion acts as a trajectory-level module rather than a special case of one editor.

\begin{table}[!t]
\centering
\caption{Compact compatibility test under the classical 700-image PIE-Bench protocol.}
\label{tab:compact_backbone}
\setlength{\tabcolsep}{3.4pt}
\renewcommand{\arraystretch}{1.04}
\scriptsize
\begin{tabular}{@{}llccc@{}}
\toprule
Backbone & Corr. & BG-LPIPS$\downarrow$ & LPIPS$\downarrow$ & CLIP-I$\uparrow$ \\
\midrule
Direct Inv. & No & 0.0633 & 0.1581 & 0.9291 \\
\oursrow Direct Inv. & Yes & \textbf{0.0584} & \textbf{0.1449} & \textbf{0.9316} \\
ProxNPI & No & 0.0786 & 0.1781 & 0.9058 \\
\oursrow ProxNPI & Yes & \textbf{0.0469} & \textbf{0.1250} & \textbf{0.9277} \\
\bottomrule
\end{tabular}
\end{table}

\subsection{Qualitative Results}

\begin{figure*}[!t]
  \centering
  \includegraphics[width=\textwidth]{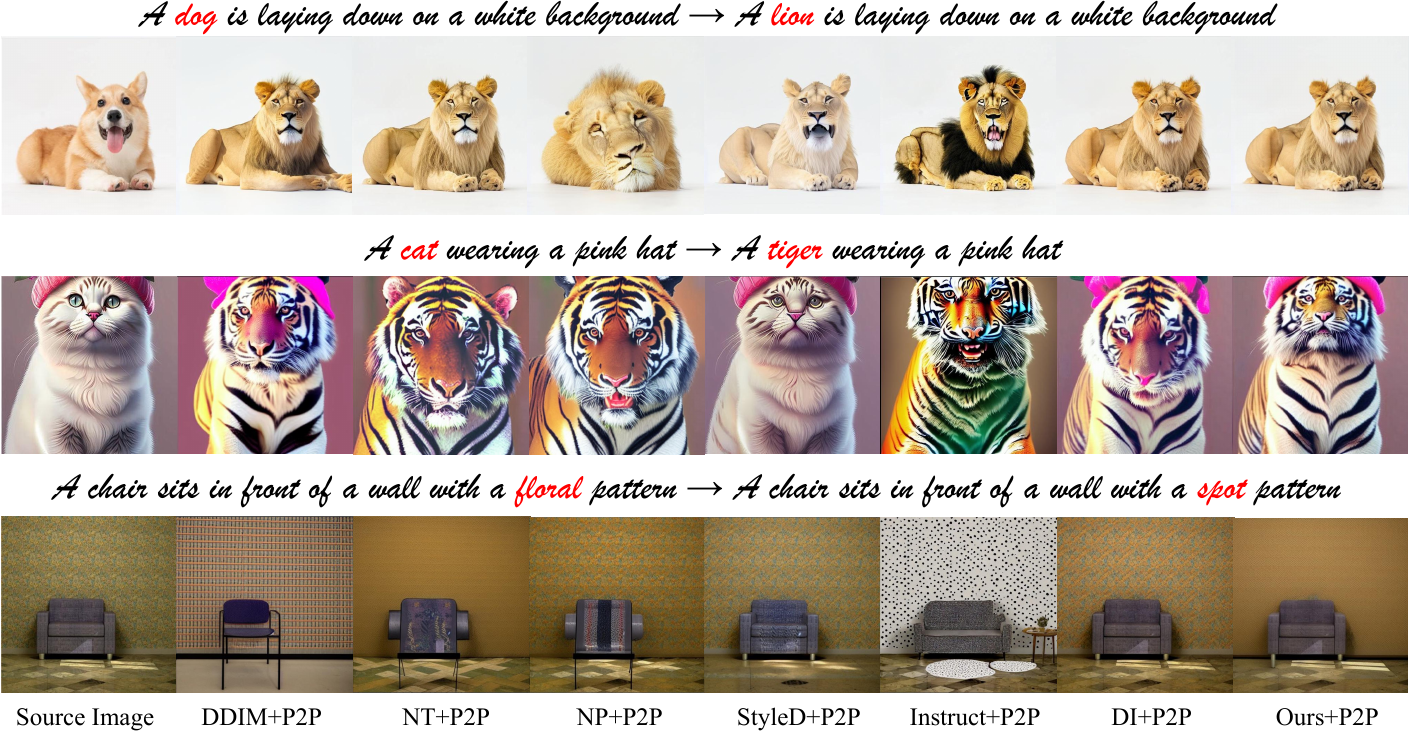}
  \caption{Qualitative comparison on prompt-based image editing. Consistent-Inversion preserves source layout and editing-irrelevant details more faithfully while keeping the target edit visible.}
  \label{fig:demo}
\end{figure*}

Fig.~\ref{fig:demo} shows representative qualitative comparisons on prompt-based image editing. Consistent-Inversion better preserves editing-irrelevant regions while allowing the target semantic change. For object replacement and attribute edits, it maintains the original pose and background layout more reliably; for texture or pattern edits, it avoids unnecessary deformation of nearby structures.

\subsection{Runtime and Reverse Consistency}
\label{sec:runtime}

Table~\ref{tab:runtime} shows that Consistent-Inversion introduces a small overhead over Direct Inversion in the classical SD-v1.4 implementation while improving reverse structural consistency. This is important for practical visual editing systems, where per-image optimization methods may be too slow for interactive use.

The runtime overhead is modest because correction is applied to only a small number of timesteps. Compared with DDIM, Consistent-Inversion increases runtime by less than one second in the current setting, while R-SSIM improves from 49.65 to 69.30. Compared with Direct Inversion, the runtime increases from 5.85s to 6.05s, but R-SSIM still improves. This supports the claim that the method provides a practical efficiency-preservation tradeoff rather than relying on expensive per-image optimization.

\begin{table}[t]
\centering
\caption{Runtime and reverse-consistency comparison.}
\label{tab:runtime}
\setlength{\tabcolsep}{5pt}
\renewcommand{\arraystretch}{1.08}
\footnotesize
\begin{tabular}{@{}lcc@{}}
\toprule
Method & Time (s)$\downarrow$ & R-SSIM$\uparrow$ \\
\midrule
DDIM & \textbf{5.16} & 49.65 \\
Direct Inversion & 5.85 & 68.86 \\
\oursrow Consistent-Inversion & 6.05 & \best{69.30} \\
\bottomrule
\end{tabular}
\end{table}

\subsection{Offset Trajectory Analysis}

We also inspect the latent offset produced by reverse consistency. In the early denoising stage, the offset magnitude is concentrated on low-frequency structure and editing-irrelevant regions, which agrees with the observation that global layout is determined before local texture refinement. After the correction is injected, the later reverse discrepancy decreases rather than merely shifting to adjacent timesteps. This indicates that the correction changes the subsequent trajectory in a persistent way, instead of acting as a transient residual perturbation.

This analysis clarifies why sparse correction is sufficient. If the offset were only a local numerical residual, correcting one or a few timesteps would have little effect after several denoising updates. In practice, early reverse consistency correction alters the source-compatible component of the target trajectory, and the following denoising steps inherit this improved geometry. This is also why the method improves preservation metrics more strongly than CLIP-T: it mainly regularizes where and how the edit evolves, not what semantic direction the target prompt requests.

\section{Discussion and Limitations}

The reverse consistency offset is useful because it is computed in the same latent space and under the same pretrained denoising network as the editing trajectory. Unlike pixel-space preservation losses or attention replacement heuristics, it estimates whether an intermediate target latent remains compatible with the source inversion path. The correction is also prompt-aware: the reverse path is conditioned on the source prompt, so the offset reflects how the pretrained diffusion model interprets the current target latent when asked to recover source semantics.

Consistent-Inversion assumes that source-to-target editing and target-to-source reversal have approximately aligned discrepancy directions during early denoising. This assumption is most suitable for edits that preserve scene layout while changing object identity, texture, material, style, or local attributes. When source and target prompts require different layouts, object counts, or viewpoints, overly strong correction may suppress the intended change. The method also adds inference overhead if correction is applied densely. These cases can be addressed by moderate correction scales, sparse timestep selection, and mask-aware schedules for complex edits.

The same principle can support more adaptive multimedia editing systems. The correction strength can depend on timestep, discrepancy magnitude, and optional editing masks, so background regions receive stronger preservation while the edited region remains flexible. Frequency-aware correction is also possible: low-frequency components can be corrected more aggressively to preserve layout, while high-frequency components remain available for texture and appearance changes. For lightweight video editing, reverse consistency offsets could be propagated across neighboring frames with optical flow or latent correspondences, encouraging source preservation within each frame and temporal stability across frames. These variants do not change the central mechanism; they expose additional controls for practical editing scenarios.

From a reproducibility perspective, the method has few moving parts: it uses the pretrained editor without finetuning and only adds a correction scale and a small set of early timesteps. The paired protocol uses the same prompts, metric code, and sampler for the base and corrected editors, isolating the effect of reverse consistency.

The quantitative evaluation is centered on PIE-Bench, which is widely used for prompt-based real-image editing and provides masks, prompt pairs, editing types, and preservation metrics. Broader user studies and stronger vision-language metrics would further clarify fine-grained instruction following and the preservation-editing tradeoff. These evaluation extensions are complementary to the central mechanism studied in this work.

\section{Conclusion}

We presented Consistent-Inversion, a training-free reverse consistency guidance framework for structure-preserving diffusion editing. By constructing an auxiliary target-side noise representation and measuring reverse consistency against the source inversion trajectory, the method obtains an interpretable correction signal for target denoising. Experiments on PIE-Bench show improved structure and background preservation with small runtime overhead, and ablations confirm the value of selected-timestep correction. The results suggest that trajectory-level reversibility is a useful principle for practical visual editing systems and can be further extended with adaptive schedules, mask-aware correction, and temporal consistency constraints.

\bibliographystyle{IEEEtran}
\bibliography{reference}

\end{document}